\documentclass[10pt, a4paper]{article}
\usepackage[final]{lrec-coling2024} 
\usepackage{listings}

\usepackage{natbib}
\usepackage{multibib}
\usepackage{multirow}
\usepackage{amsmath}
\usepackage{amssymb}
\usepackage{comment}
\usepackage{booktabs}
\usepackage[ruled,vlined]{algorithm2e}

\makeatletter
\def\@mb@citenamelist{cite,citep,citet,citealp,citealt,citepalias,citetalias}
\makeatother
\newcites{languageresource}{~}

\usepackage{graphicx}
\usepackage{tabularx}
\usepackage{soul}
\usepackage{xcolor}
\usepackage{hyperref}
 \definecolor{darkblue}{rgb}{0, 0, 0.5}
  \hypersetup{colorlinks=true, citecolor=darkblue, linkcolor=darkblue, urlcolor=darkblue}
\usepackage{xstring}
\usepackage{color}
\usepackage{caption}

\title{Sinkhorn Distance Minimization for Knowledge Distillation}
\name{Xiao Cui$^1$, Yulei Qin$^{2*}$\thanks{$^*$Contribute equally with the first author.}, Yuting Gao$^2$, Enwei Zhang$^2$, Zihan Xu$^2$, Tong Wu$^2$, \\{\bf \large Ke Li$^2$, Xing Sun$^2$, Wengang Zhou$^{1\dagger}$ and Houqiang Li$^{1\dagger}$\thanks{$^{\dagger}$Corresponding authors: Wengang Zhou and Houqiang Li.}}
} 

\address{$^1$University of Science and Technology of China, $^2$Tencent YouTu Lab
 \\
         cuixiao@mail.ustc.edu.cn, 
         \{zhwg,lihq\}@ustc.edu.cn, \\
         \{yuleiqin, yutinggao, miyozhang, ianxxu, townswu,   tristanli, winfredsun\}@tencent.com\\}

\abstract{
Knowledge distillation (KD) has been widely adopted to compress large language models (LLMs).
Existing KD methods investigate various divergence measures including the Kullback-Leibler (KL),
reverse Kullback-Leibler (RKL),
and Jensen-Shannon (JS) divergences.
However,
due to limitations inherent in their assumptions and definitions,
these measures fail to deliver effective supervision when few distribution overlap exists between the teacher and the student.
In this paper,
we show that the aforementioned KL, RKL, and JS divergences respectively suffer from issues of mode-averaging, mode-collapsing, and mode-underestimation,
which deteriorates logits-based KD for diverse NLP tasks.
We propose the Sinkhorn Knowledge Distillation (SinKD) that exploits the Sinkhorn distance to ensure a nuanced and precise assessment of the disparity between teacher and student distributions.
Besides,
profit by properties of the Sinkhorn metric,
we can get rid of sample-wise KD that restricts the perception of divergence in each teacher-student sample pair.
Instead,
we propose a batch-wise reformulation to capture geometric intricacies of distributions across samples in the high-dimensional space.
Comprehensive evaluation on GLUE and SuperGLUE,
in terms of comparability, validity, and generalizability,
highlights our superiority over state-of-the-art methods on all kinds of LLMs with encoder-only, encoder-decoder, and decoder-only architectures.
 \\ \newline  
}

\begin{document}

\maketitleabstract

\section{Introduction}
Large language models (LLMs) such as BERT~\cite{bert},
RoBERTa~\cite{roberta},
T0~\cite{t0},
and GPT~\cite{gpt2,gpt3}
have set state-of-the-art (SOTA) records on various tasks in the field of natural language processing (NLP).
On one hand,
the scaling laws of LLMs undoubtedly stimulate the development of models with billions of parameters.
On the other hand,
the surge of model size makes it unaffordable for LLMs to be deployed under resource-constrained environments.
Consequently,
knowledge distillation (KD),
emerging as a cost-efficient approach,
has attracted attention from researchers to distill smaller models which maintain highly competitive performance.

\begin{figure}[!ht]
\begin{center}
\includegraphics[width=1\linewidth]{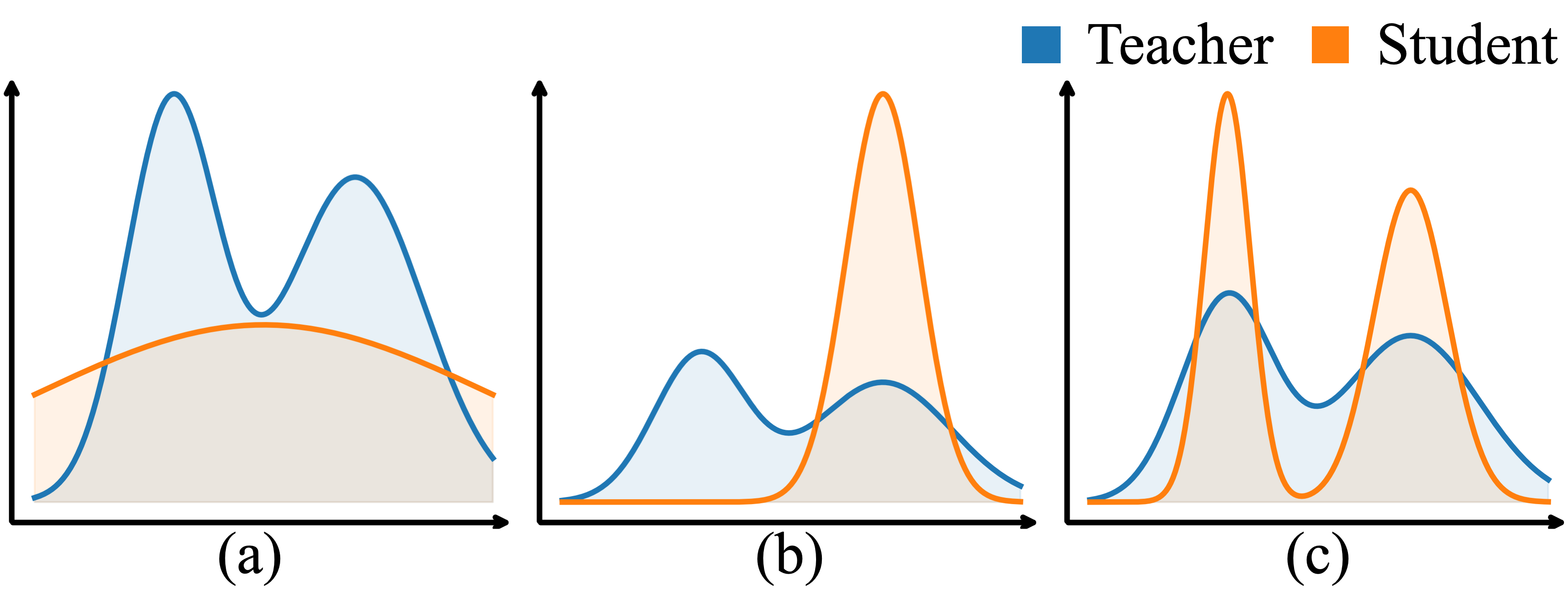}
\caption{
Limitations of existing divergence measures for the student to match the teacher in logits-based distillation.
(a) Mode-averaging by Kullback-Leibler divergence.
(b) Mode-collapsing by reverse Kullback-Leibler divergence.
(c) Mode-underestimation by Jensen-Shannon divergence.
}
\label{fengmian}
\end{center}
\end{figure}

One kind of the most representative KD methods is logits-based KD,
where the divergence between the distributions of the predicted logits from teacher and student models is measured and minimized for knowledge transfer.
The key to effective logits-based KD is exactly the proper measurement of such divergence.
Existing studies have experimented with Kullback-Leibler (KL) divergence~\cite{kd},
reverse Kullback-Leibler (RKL) divergence~\cite{rkl,minillm},
and Jensen-Shannon (JS) divergence~\cite{f-div,js1,js2}.
All these measures can be viewed as variants of the $f$-divergence measures,
which are notoriously limited in quantification of distributions that lack substantial intersections~\cite{wgan1}.
Moreover,
as illustrated in Fig.~\ref{fengmian},
each measure has its own drawbacks.
KL distillation results in a \textbf{mode-averaging} issue~\cite{kim2016sequence, ijcai2021p362},
causing the student to learn an excessively smooth distribution that encompasses the entire support of the teacher distribution.
RKL distillation leads to \textbf{mode-collapsing}~\cite{arjovsky2017towards, f-div},
where the student overly focuses on one of the highly probable regions of the teacher distribution and ignores the remaining one.
JS distillation gives rise to \textbf{mode-underestimation}~\cite{nowozin2016f, yu2020training}
where the student underestimates the probability of rare events due to insufficient penalty.

Another challenge of performing sample-wise KD on LLMs is that for discriminative tasks,
the low-dimensional categorical outputs from the teacher provide
limited insights into their underlying distributions in the high-dimensional hidden space.
One intuitive solution is to bring in a batch of samples to collectively grasp the distribution differences.
Nevertheless,
existing divergence measures can only independently deal with each sample for logit-by-logit matching because they are not distance metrics and cannot locate the paired teacher and student logits of the same sample from the batch for overall distance minimization.

To address these challenges,
we propose \textbf{Sin}khorn \textbf{K}nowledge \textbf{D}istillation,
termed as SinKD,
for distillation of LLMs\footnote{Codes and models are available at
\url{https://github.com/2018cx/SinKD}.}.
In consideration of generalizability,
we tackle logits-based KD in the present study,
which would benefit a broad range of applications.
Our SinKD employs the Sinkhorn distance~\cite{sinkhornbase},
a variant of the Wasserstein distance~\cite{was2, was},
as divergence measure.
The Wasserstein distance quantifies the dissimilarity between two distributions by calculating the minimum cost required to transform one distribution into the other.
Compared with traditional divergence measures,
it is more sensible as a cost function for distillation.
Furthermore,
it is differentiable almost everywhere,
enabling easy optimization.
Despite these advantages,
the Wasserstein distance itself is difficult to be computed analytically.
Its associated computational cost is prohibitively high for distilling LLMs.
Under such circumstance,
we propose to use Sinkhorn distance as an efficient approximation,
which not only retains all the benefits of Wasserstein distance but also greatly mitigates its cost issue.

A straightforward application of Sinkhorn distance on sample-wise logits matching,
though feasible,
cannot take full advantage of its perception of structural differences in distributions.
Fortunately,
Sinkhorn distance is a symmetric metric and its derivation from the optimal transport (OT) imposes explicit constraints on \textbf{matching correctness}.
It means that given a batch of logits outputs from the teacher and the student respectively as sets A and B,
the minimization of the overall Sinkhorn distance between A and B enforces a precise element-wise matching between the two outputs coming from the same sample in a batch.
Such properties allow it to work beyond sample-wise distillation.
As a result,
we propose
the batch-wise reformulation.
In this way,
we can capture geometric structures of the intricate and implicit distributions even through low-dimensional observations.
We do not introduce additional network layer or modify output formats specific to NLP tasks.

Extensive experiments are conducted in view of 1) \textbf{comparability},
2) \textbf{validity},
and 3) \textbf{generalizability}.
For comparability,
we test SinKD with BERT on the GLUE benchmark~\cite{glue} and it consistently outperforms the SOTA KD methods.
For validity,
we provide a comprehensive analysis on ablation studies and hyper-parameters,
Our findings
advise practitioners on
how to adopt SinKD in their own work.
For generalizability,
we test SinKD on the SuperGLUE benchmark~\cite{superglue} with LLMs of various architectures,
ranging from the encoder-decoder T0~\cite{t0} to the decoder-only GPT-Neo~\cite{gpt-neo} transformers.
Our SinKD showcases robustness across model choices while previous studies merely investigate KD techniques on the encoder-only BERT.

In summary, our contributions are:
\begin{itemize}
    \item We propose SinKD,
    a knowledge distillation approach that employs the Sinkhorn distance for divergence measurement.
    It not only addresses limitations of KL, RKL, and JS divergences under extreme distribution scenarios,
    but also circumvents the computation burden of Wasserstein distance for distillation.
    \item We unearth the properties of Sinkhorn distance and reformulate SinKD
    into batch-wise OT,
    extending its applicability in NLP tasks.
    \item Extensive experiments in terms of comparability, validity, and generalizability demonstrate the superiority of SinKD over SOTA methods.
    We offer practical guidelines of distilling various LLMs
    for real-world applications.
\end{itemize}



\section{Related Work}
\subsection{Knowledge Distillation}

Knowledge distillation is initially introduced by~\cite{2006} where an ensemble of models act as the teacher to train a single student model,
and now frequently referred to as a model compression technique.
Existing KD methods can be simply classified into two categories:
1) logits-based KD and
2) representation-based KD.
The logits-based KD is popularized by~\cite{kd}.
They force the student to match the predictions of the teacher as soft targets via cross-entropy loss,
which is equivalent to minimize the KL divergence between teacher and student probabilities.
\cite{kim2016sequence} bring logits-based KD into generative language models and propose sequence KD.
\cite{distilbert} and~\cite{pd} apply KD on BERT for smaller models with minor degradation.
\cite{rkl} propose ENGINE to use the reverse KL for distillation of a non-autoregressive translation model.
For representation-based KD,
the hidden, intermediate representations of input tokens have been utilized as the matching targets of the student~\cite{tinybert,distilbert,adkd}.
There also exist methods that can adapt to either logits-based or representation-based KD~\cite{metadistill,reaugkd}.

In this paper,
we primarily focus on logits-based KD and investigate the fundamental problem:
\textit{how to transfer label-supplementary knowledge from the teacher to the student with an effective and reliable divergence measure}.
Previous studies exploit KL divergence~\cite{kd},
RKL divergence~\cite{rkl,minillm},
JS divergence~\cite{f-div,js1,js2},
and sophisticated distance measures~\cite{pkd,rco,rkd,sftn} for distillation.
However,
these methods do not consistently capture subtle distribution differences
and tend to take "shortcuts" in student imitating the teacher,
which motivates our exploration of an alternative divergence measure.

\subsection{Sinkhorn Distance}

We first introduce the Wasserstein distance 
as a foundation for the Sinkhorn distance.
It is a dissimilarity metric
derived by the mass transportation theory of two probability measures.
Since the Wasserstein distance takes into account the underlying geometry of the
distribution space~\cite{was,was2,wcal,wd},
it enjoys high popularity in
generative adversarial networks~\cite{wgan1,wgan2,wgan3} and unsupervised learning~\cite{un1,un2,un3}.
However,
the Wasserstein distance is too costly to be computed and its efficient approximation is a prerequisite for distillation.
The Sinkhorn distance stems from it
and incorporates an extra entropy regularization term to make the OT tractable.
It is informally defined by the minimum transport cost of an entropy-regularized OT plan~\cite{sinkhornbase},
and has been successful in classification~\cite{frogner2015,liu2023bilaterally},
machine translation~\cite{li2023improving},
domain adaptation~\cite{courty2017,nguyen2022improving},
and generative modeling~\cite{genevay2018,kammammettu2023scenario}.

For distillation of LLMs,
especially under discriminative tasks,
the vanilla sample-wise SinKD
cannot make the best use of its desirable properties in perceiving structural differences between distributions.
On the contrary,
we propose the batch-wise SinKD to make up the insufficient knowledge revealed from the low-dimensional outputs of the teacher,
improving its generalization over tasks.



\section{Methodology}

In this section,
we first review classic divergence
measures and analyze their drawbacks.
Then,
we present details of SinKD with an OT framework.

\subsection{Problem Statement}
Given a sample $\mathbf{x}_{i}$ and its ground-truth label $\mathbf{y}_{i}\in\mathbb{R}^{d}$ in the training set,
the output logits with softmax activation $\sigma_{\tau}$ from the teacher $f_T$ and the student $f_S$ are respectively $\mathbf{t}_{i}\in\mathbb{R}^{d}$ and $\mathbf{s}_{i}\in\mathbb{R}^{d}$:
\begin{equation}
    \mathbf{t}_{i}=\sigma_{\tau}(f_T (\mathbf{x}_i)),\quad \mathbf{s}_{i}=\sigma_{\tau}(f_S (\mathbf{x}_i)),
    \label{eq:3}
\end{equation}
where $\tau$ is the temperature and $d$ is the dimension of the output logits.
The objective of KD is to minimize the measured divergence $J(\mathbf{t}_{i},\mathbf{s}_{i})$.

\subsection{Classic Divergence Measures}

\paragraph{KL Divergence}

It quantifies the amount of information lost when 
$\mathbf{s}_{i}$ approximates $\mathbf{t}_{i}$ as:
\begin{equation}
J_{\text{KL}}(\mathbf{t}_{i},\mathbf{s}_{i})\thickapprox\sum_{j=1}^{d}(-{\mathbf{t}_{i(j)}\log{\mathbf{s}_{i(j)}}+\mathbf{t}_{i(j)}\log{\mathbf{t}_{i(j)}}}).
\end{equation}
Here,
$j$ denotes the index of an element in a vector.
Despite its popularity,
KL divergence suffers from three limitations.
First,
it is asymmetric with
$J_{\text{KL}}(\mathbf{t}_{i},\mathbf{s}_{i})\neq J_{\text{KL}}(\mathbf{s}_{i},\mathbf{t}_{i})$,
which introduces inconsistencies due to its violation of the property as a distance metric.
Second,
the student model optimized by the KL loss
attempts to average the teacher's multimodal distribution,
ending up with an underfitting of these modes.
This is known as the mode-averaging problem.
Consequently,
the student fails to capture all crucial patterns of data and ultimately impacts performance.
Third,
the KL divergence corresponds to a non-smooth function,
posing challenges to model optimization.



\paragraph{RKL Divergence} It addresses the issue of mode-averaging associated with $J_{\text{KL}}(\mathbf{t}_{i},\mathbf{s}_{i})$:
\begin{equation}
J_{\text{RKL}}(\mathbf{t}_{i},\mathbf{s}_{i})\thickapprox\sum_{j=1}^{d}({\mathbf{s}_{i(j)}\log{\mathbf{s}_{i(j)}}}-{\mathbf{s}_{i(j)}\log{\mathbf{t}_{i(j)}}}).
\end{equation}
However,
it shares the inherent asymmetry with KL which leads to inconsistencies in capturing differences.
Furthermore,
the student optimized by a RKL loss tends to pay attention only to highly likely events of the teacher's distribution,
which is known as mode-collapsing.
Accordingly,
if the teacher assigns zero-probability to an event,
the student is compelled to do the same.
This ``zero-forcing" effect could be problematic as the student lacks the capacity to track the complete distribution of the teacher,
resulting in suboptimal performance.


\paragraph{JS Divergence} It combines both KL and RKL by:
\begin{equation}
\begin{split}
J_{\text{JS}}(\mathbf{t}_{i},\mathbf{s}_{i})&\thickapprox\frac{1}{2}\sum_{j=1}^{d}(-{\mathbf{s}_{i(j)}\log{\mathbf{m}_{i(j)}}+\mathbf{s}_{i(j)}\log{\mathbf{s}_{i(j)}}}\\
&-{\mathbf{t}_{i(j)}\log{\mathbf{m}_{i(j)}}+\mathbf{t}_{i(j)}\log{\mathbf{t}_{i(j)}}}),
\end{split}
\end{equation}
where $\mathbf{m}_{i}=\frac{1}{2}(\mathbf{t}_{i}+\mathbf{s}_{i})$.
While the JS divergence overcomes the asymmetry shortcoming of the KL divergence,
it is still subject to non-smoothness that makes it challenging to optimize.
Moreoever,
the student may excessively underestimate the probability of rare events as the JS loss does not penalize adequately for matching low probability regions.
There also exists a risk of gradient vanishing when $J_{\text{JS}}(\mathbf{t}_{i},\mathbf{s}_{i})$ degenerates as a constant on distributions with few or even no overlap.

\begin{figure}[t]
\begin{center}
\includegraphics[width=1\linewidth]{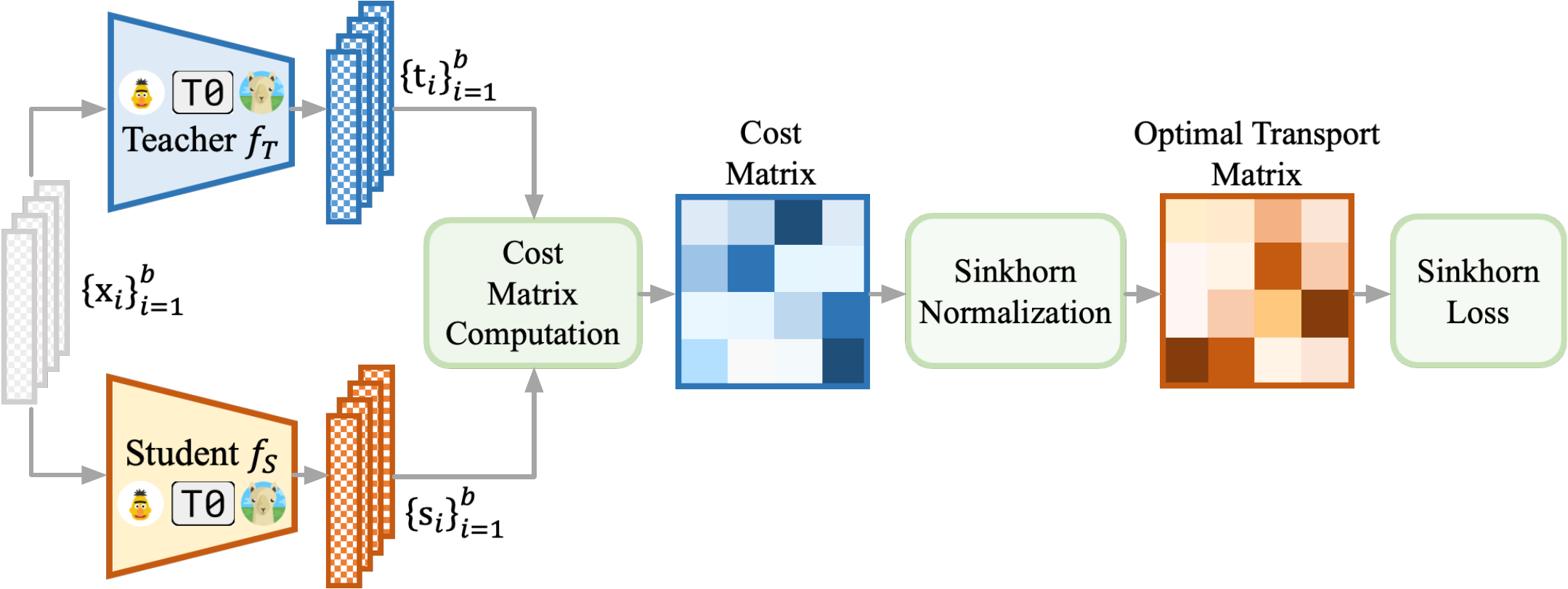}
\caption{Illustration of our SinKD pipeline.
}
\label{method}
\end{center}
\end{figure}

\subsection{Sinkhorn Distance}
Sinkhorn distance
is based on the relaxed formulation of an OT plan with entropy regularization.
It considers the minimum cost of mass transmission in converting one probability into the other.
Specifically,
we first define the Wasserstein distance below.
It involves the set of a transportation polytope $U(\mathbf{t}_{i},\mathbf{s}_{i})$,
which consists of all matrices of $\mathbf{P}\in \mathbb{R}_{+}^{d\times d}$ that satisfy the following constraints:
\begin{equation}
U(\mathbf{t}_{i},\mathbf{s}_{i})=\{\mathbf{P}\in \mathbb{R}_{+}^{d\times d}|\mathbf{P1}_d = \mathbf{s}_{i},\mathbf{P}^{\text{T}}\mathbf{1}_d = \mathbf{t}_{i}\},
\end{equation}
where $\mathbf{1}_d\in\mathbb{R}^{d}$ is a vector of ones.
Given a cost matrix $\mathbf{D}\in\mathbb{R}^{d\times d}$,
the Wasserstein distance is:
\begin{equation}
J_\text{WD}(\mathbf{t}_{i},\mathbf{s}_{i}) = \underset{\mathbf{P}\in U(\mathbf{t}_{i},\mathbf{s}_{i})}{\min}\left<\mathbf{P},\mathbf{D}\right>=\sum_{m,n}{\mathbf{P}_{m,n}\mathbf{D}_{m,n}},
\end{equation}
where $\mathbf{D}_{m,n}$ is usually the absolute difference between the $m$-th and $n$-th elements of $\mathbf{t}_{i}$ and $\mathbf{s}_{i}$:
\begin{equation}
    \mathbf{D}_{m,n} = \vert\mathbf{t}_{i(m)} - \mathbf{s}_{i(n)}\vert.
\end{equation}
To circumvent the substantial computation entailed by solving such an OT problem,
Sinkhorn distance is proposed as a fast approximation to the Wasserstein distance for a constrained optimization~\cite{sinkhornbase}.
It is defined as the inner product between the OT plan $\mathbf{P}^{\lambda}$ and the cost matrix $\mathbf{D}$:
\begin{equation}
J_{\text{SD}}(\mathbf{t}_{i},\mathbf{s}_{i})=\left<\mathbf{P}^{\lambda},\mathbf{D}\right>,
\end{equation}
where $\lambda>0$ is the weight for entropy regularization.
The OT plan $\mathbf{P}^{\lambda}$ is obtained by minimizing:
\begin{equation}
    \mathbf{P}^{\lambda}= \underset{\mathbf{P}\in U(\mathbf{t}_{i},\mathbf{s}_{i})}{\text{argmin}}\left<\mathbf{P},\mathbf{D}\right>-\lambda h(\mathbf{P}),
\end{equation}
where $h(\mathbf{P})$ is the entropy of the matrix $\mathbf{P}$.
The entropy term encourages the transport plan to be more spread out for easier optimization.
The vanilla solution to $\mathbf{P}^{\lambda}$ by sample-wise Sinkhorn normalization~\cite{sinkhornbase} is performed between $\mathbf{t}_i$ and $\mathbf{s}_i$ in a manner of iterative updates:
\begin{equation}
\left(\mathbf{u}^t,\mathbf{v}^t\right)\leftarrow \left(\mathbf{t}_i\oslash\left(\mathbf{K}^{\text{T}}\mathbf{v}^{t-1}\right),\mathbf{s}_i\oslash\left(\mathbf{K}\mathbf{u}^{t-1}\right)\right),
\label{eq:solution2P}
\end{equation}
where $\oslash$ indicates element-wise division and $t$ denotes the iteration time.
Two vectors $\mathbf{u}\in\mathbb{R}^{d},\mathbf{v}\in\mathbb{R}^{d}$ are non-negative.
The kernel matrix $\mathbf{K}\in\mathbb{R}^{d\times d}$ is constructed by applying the Gaussian kernel on $\mathbf{D}$ with the weight $\lambda$ for entropy regularization:
\begin{equation}
\label{eq:initK}
    \mathbf{K}=\exp(-\frac{\mathbf{D}}{\lambda}).
\end{equation}
Finally,
$\mathbf{P}^{\lambda}$ is defined as:
\begin{equation}
\mathbf{P}^{\lambda}=\mathrm{diag} \left(\mathbf{v}^t\right)\mathbf{K}\mathrm{diag} \left(\mathbf{u}^t\right).
\end{equation}

\subsection{Batch-wise Reformulation}

In view of properties of the Sinkhorn distance metric,
we can get rid of the sample-wise KD that only works on each teacher-student sample pair,
and instead perform KD on groups of teacher and student samples.
A batch of $b$ samples all participate in divergence measures with their overall output logits $\mathbf{t}\in\mathbb{R}^{b\times d}$ and $\mathbf{s}\in\mathbb{R}^{b\times d}$ respectively from the teacher and the student.
It thereby increases the dimension of the ``observational" space via batch-wise reformulation especially when $d\ll b$ holds.

\paragraph{Cost Matrix Computation} We employ the $\ell_p$-norm to measure the pairwise differences between the $i$-th and $j$-th samples in a batch for the entry $\mathbf{D}_{i,j}$ of the ``batchified" cost matrix $\mathbf{D}\in\mathbb{R}^{b\times b}$:
\begin{equation}
\label{eq:batchcost}
    \mathbf{D}_{i,j} = \lVert\mathbf{t}_{i} - \mathbf{s}_{j}\rVert_p.
\end{equation}

\paragraph{Sinkhorn Normalization}
Before we propose the batch-wise Sinkhorn normalization,
we reformulate the sample-wise solution to $\mathbf{P}^{\lambda}$ (Eq.~\ref{eq:solution2P}) into a equivalent vector-form with iterations only on $\mathbf{K}$:
\begin{equation}
\begin{aligned}
&\mathbf{\widehat{K}}^{t} \leftarrow \mathrm{diag} \left(\mathbf{K}^{t-1}\mathbf{1}_{d}\oslash\mathbf{s}_i\right)^{-1}\mathbf{K}^{t-1}, \\
&\mathbf{K}^{t} \leftarrow \mathbf{\widehat{K}}^{t}\mathrm{diag} \left(\left(\mathbf{\widehat{K}}^{t}\right)^{\text{T}}\mathbf{1}_{d}\oslash\mathbf{t}_i\right)^{-1},
\end{aligned}
\end{equation}
where $\mathbf{K}^{0}=\mathbf{K}\in\mathbb{R}^{d\times d}$ is defined in Eq.~\ref{eq:initK}.
For distillation beyond the $d$-dimensional space,
we propose a more compact solution in the matrix-form for batch-wise normalization with $\mathbf{K}\in\mathbb{R}^{b\times b}$:
\begin{equation}
\label{eq:updateK}
\begin{aligned}
&\mathbf{\widehat{K}}^{t} \leftarrow \mathrm{diag} \left(\mathbf{K}^{t-1}\mathbf{1}_{b}\oslash\mathbf{w}_s\right)^{-1}\mathbf{K}^{t-1}, \\
&\mathbf{K}^{t} \leftarrow \mathbf{\widehat{K}}^{t}\mathrm{diag} \left(\left(\mathbf{\widehat{K}}^{t}\right)^{\text{T}}\mathbf{1}_{b}\oslash\mathbf{w}_t\right)^{-1},
\end{aligned}
\end{equation}
where $ \mathbf{w}_s$ and $\mathbf{w}_t$ respectively represent the weights of each element involved in the batch-wise KD from the student and teacher.
Without loss of generality,
we assume uniform distributions with $ \mathbf{w}_s=\mathbf{w}_t=\frac{1}{b}\mathbf{1}_{b}$.
Given such conditions,
updates on $\mathbf{K}^{t}$ (Eq.~\ref{eq:updateK}) can be further simiplified as:
\begin{equation}
\label{eq:updateK2}
\begin{aligned}
\mathbf{\widehat{K}}^{t} &\leftarrow \mathbf{K}^{t-1} \oslash \left( \mathbf{K}^{t-1}\mathbf{1}_{b}\mathbf{1}_{b}^\top \right), \\
\mathbf{K}^t &\leftarrow \mathbf{\widehat{K}}^{t} \oslash \left( \mathbf{1}_{b}\mathbf{1}_{b}^\top\mathbf{\widehat{K}}^{t} \right).
\end{aligned}
\end{equation} 
Out of simplicity,
irrelevant constants are excluded from the equations above.
With a pre-determined number of iterations $T$, the OT matrix is derived:
\begin{equation}
    \mathbf{P}^{\lambda}=\mathbf{K}^{T}
\end{equation}

\paragraph{Sinkhorn Loss} The batch-wise SinKD loss is: 
\begin{equation}
\mathcal{L}_{\text{SD}}=J_{\text{SD}}(\mathbf{t},\mathbf{s})=\left<\mathbf{P}^{\lambda},\mathbf{D}\right>=\sum_{i,j}{\mathbf{K}^{T}_{i,j}\mathbf{D}_{i,j}}
\end{equation}
We illustrate the entire pipeline in Fig.~\ref{method}.

\paragraph{Total Losses}
For each batch of $b$ samples,
we use the cross-entropy loss $\mathcal{L}_{\text{CE}}$,
the KL loss $\mathcal{L}_{\text{KL}}$,
and the Sinkhorn loss $\mathcal{L}_{\text{SD}}$ for distillation:
\begin{equation}
\begin{split}
    \mathcal{L}&=\sum_{i=1}^{b}[(1-\alpha)\mathcal{L}_{\text{CE}}(\mathbf{y}_{i},\mathbf{s}_{i})\\
    &+\alpha\mathcal{L}_{\text{KL}}(\mathbf{t}_{i},\mathbf{s}_{i})]+\beta\mathcal{L}_{\text{SD}},
\end{split}
\end{equation}
where $\alpha$ and $\beta$ are weights,
and $\mathcal{L}_{\text{KL}}(\mathbf{t}_{i},\mathbf{s}_{i})\thickapprox\mathcal{L}_{\text{CE}}(\mathbf{t}_{i},\mathbf{s}_{i})$ given that the second term in $J_{\text{KL}}(\mathbf{t}_{i},\mathbf{s}_{i})$ can be viewed as a constant in distillation.

\paragraph{Alternative $\mathbf{D}$}
Apart from Eq.~\ref{eq:batchcost},
we can further take into account all the $d$-dimensional logits of $b$ samples by flattenning $\mathbf{t}$ and $\mathbf{s}$ for a $\mathbf{D}\in\mathbb{R}^{bd\times bd}$:
\begin{equation}
\label{eq:batchcost2}
    \mathbf{D}_{im,jn} = \vert\mathbf{t}_{i(m)} - \mathbf{s}_{j(n)}\vert.
\end{equation}
Accordingly,
the sinkhorn normalization is performed on $\mathbf{K}\in\mathbb{R}^{bd\times bd}$ with $\mathbf{w}_s=\mathbf{w}_t=\frac{1}{bd}\mathbf{1}_{bd}$.
In this case,
SinKD takes a broader perspective of the batch distributions with a multiplied dimension of $bd$,
significantly exceeding the sample-wise KD.


\section{Experimental Settings}

\subsection{Datasets}
We evaluate our method on seven tasks of the GLUE benchmark~\cite{glue},
including CoLA~\cite{cola},
SST-2~\cite{sst},
MNLI~\cite{mnli},
MRPC~\cite{mrpc},
RTE~\cite{rte},
QNLI~\cite{qnli} and QQP~\cite{qqp}.
For evaluation metrics,
we follow previous
works~\cite{adkd,reaugkd,metadistill} to report accuracy (MNLI, SST-2, QNLI, QQP, and RTE),
F1 score (MRPC),
and Matthews correlation coefficient (CoLA).
Following~\cite{reaugkd,metadistill},
the regression-oriented STS-B~\cite{sts} is not validated due to its problem settings.
Note that all discriminative tasks of GLUE are associated with extremely-low dimension of logits output ($d=3$ for MNLI and $d=2$ for the remainings tasks).

\subsection{Implementation Details}

Our SinKD is implemented with PyTorch
and Transformers~\cite{transformers}.
For comparability,
we follow AD-KD~\cite{adkd} to set BERT$_{\text{base}}$ as the teacher and a smaller BERT$_{\text{6}}$~\cite{pd} as the student for task-specific fine-tuning. 
For generalizability,
we also validate SinKD on T0~\cite{t0} and GPT-Neo~\cite{gpt-neo}.
Note that for all GLUE tasks except MNLI,
two definitions of $\mathbf{D}$ (Eqs.~\ref{eq:batchcost},\ref{eq:batchcost2}) are equivalent given the constraint of $\sum_{m=1}^{d}\mathbf{t}_{i(m)}=1$ and $d=2$.
Consequently,
we use the default $\mathbf{D}$ by Eq.~\ref{eq:batchcost}.
Out of simplicity,
we set $p=1$ ($\ell_1$-norm) for $\mathbf{D}$.
The hyper-parameters are optimized via grid search to determine the learning rate $lr\in\{2e-5, 3e-5, 4e-5, 5e-5\}$,
$\alpha\in\{0.8, 0.9, 1.0\}$,
$b\in\{16, 32, 64\}$,
and $\tau_{\text{KL}}
\in\{1, 2, 3, 4\}$.
We empirically set
$\tau_{\text{SD}}=2$,
$\lambda=0.1$,
$T=20$,
and $\beta=0.8$.
Discussions on the effect of $T$,
$\lambda$,
$\tau_{\text{SD}}$,
$\alpha$,
and $\beta$ can be found in Sec.~\ref{5.3}.

\subsection{Baselines}

We compare SinKD with SOTA KD methods on logits and representations.
For logits-based KD,
we include the vanilla KD~\cite{kd},
RCO~\cite{rco},
DML~\cite{dml},
PD~\cite{pd},
and ReAugKD~\cite{reaugkd}.
For representation-based KD,
we compare PKD~\cite{pkd},
TinyBERT~\cite{tinybert},
RKD~\cite{rkd},
CKD~\cite{ckd},
SFTN~\cite{sftn},
TAKD~\cite{takd},
ProKT~\cite{prokt},
MGSKD~\cite{prokt},
MetaDistill~\cite{metadistill},
and AD-KD~\cite{adkd}.
For a fair comparison,
we follow~\cite{adkd} to exclude MiniLM~\cite{minilm} and MobileBERT~\cite{mobilebert} as their two-stage settings involve both task-agnostic and task-specific distillation.
In contrast,
we emphasize a more generalized one-stage
setting where no extra efforts are required for
pre-training.
Baseline results are quoted~\cite{adkd,reaugkd}.


\section{Results and Discussions}

\begin{table*}[t]
  \centering
  \resizebox{1\linewidth}{!}{
    \begin{tabular}{l|c|ccccccc|c}
    \toprule
    \multirow{2}{*}{\textbf{Method}}
    & \multirow{2}{*}{\textbf{\#Params.}} & \textbf{COLA} & \textbf{SST-2} & \textbf{MNLI-(m/mm)} & \textbf{MRPC} & \textbf{RTE} & \textbf{QNLI} & \textbf{QQP} & \multirow{2}{*}{\textbf{Avg}}\\ 
    & & (MCC) & (ACC) & (ACC) & (F1) & (ACC) & (ACC) & (ACC) & \\
    \midrule
    {BERT}$_{\text{base}}$ (T)~\cite{bert} & 110M & 60.3 & 93.7 & 84.9/84.8 & 91.4 & 71.1 & 91.7 & 91.5 &83.28  \\
    {BERT}$_{\text{6}}$ (S)~\cite{pd} & 66M & 51.2 & 91.0 & 81.7/82.6 & 89.2 & 66.1 & 89.3 & 90.4 &79.53 \\
    \midrule
    \multicolumn{10}{c}{Task-specific Representation-based Distillation} \\
    \midrule
    PKD~\cite{pkd} & 66M & 45.5 & 91.3 & 81.3/- & 85.7 & 66.5 & 88.4 & 88.4 &77.63  \\
    TinyBERT~\cite{tinybert} & 66M & 53.8 & 92.3 & 83.1/83.4 & 88.8 & 66.5 & 89.9 & 90.5 &80.3 \\
    RKD~\cite{rkd} & 66M & 53.4 & 91.7 & - & 86.1 & 68.6 & 89.5 & 90.9 & 80.03  \\
    CKD~\cite{ckd} & 66M & 55.1 & 93.0 & 83.6/84.1 & 89.6 & 67.3 & 90.5 & 91.2 &81.11 \\
    SFTN~\cite{sftn}& 66M & 53.6 & 91.5 & - & 85.3 & 68.5 & 89.5 & 90.4 & 79.80  \\
    TAKD~\cite{takd} & 66M & 53.8 & 91.4 & - & 85.0 & 68.5 & 89.6 & 90.7 & 79.83  \\
    ProKT~\cite{prokt} & 66M & 54.3 & 91.3 & - & 86.3 & 68.4 & 89.7 & 90.9 & 80.15  \\
    MGSKD~\cite{mgskd} & 66M & 49.1 & 91.7 & 83.3/83.9 & 89.8 & 67.9 & 90.3 & 91.2 &80.00 \\
    MetaDistill~\cite{metadistill} & 66M & 58.6 & 92.3 & - & 86.8 & 69.4 & 90.4 & 91.0 & 81.42  \\
    AD-KD~\cite{adkd} & 66M & 58.3 & 91.9 & 83.4/\textbf{84.2} & 91.2 & 70.9 & \textbf{91.2} & 91.2 &82.45 \\
    \midrule
    \multicolumn{10}{c}{Task-specific Logits-based Distillation} \\
    \midrule
    Vanilla KD~\cite{kd}   & 66M & 53.6 & 91.1 & 82.7/83.1 & 89.4 & 66.8 & 90.1 & 90.5 &80.25 \\
    RCO~\cite{rco} & 66M & 53.6 & 91.4 & - & 85.1 & 67.6 & 89.7 & 90.6 & 79.67  \\
    DML~\cite{dml} & 66M & 53.7 & 91.5 & - & 85.1 & 68.4 & 89.6 & 90.3 & 79.77  \\
    PD~\cite{pd} & 66M & - & 91.1 & 82.5/83.4 & 89.4 & 66.7 & 89.4 & 90.7 & -\\
    ReAugKD~\cite{reaugkd} & 66M & 59.4 & 92.5 & - & 86.3 & 70.4 & 90.7 & 91.2 & 81.75  \\
    \midrule
    SinKD (ours)  &66M & \textbf{60.2} & \textbf{93.1} & \textbf{83.8}/\textbf{84.2} & \textbf{91.3} & \textbf{71.1} & 90.5 & \textbf{91.3} &\textbf{82.92} \\
    \bottomrule
    \end{tabular}}%
    \caption{Comparison with SOTA methods on GLUE with {BERT}$_{\text{base}}$ as the teacher (T) and {BERT}$_{\text{6}}$ as the student (S). All scores are averaged except the accuracy of MNLI-(m/mm).
    }
  \label{sota}%
\end{table*}%

\subsection{Comparison with SOTA}
Tab.~\ref{sota}
shows that SinKD outperforms all baselines on most datasets.
Specifically,
SinKD achieves an average increase of 0.47\% and 1.17\% over AD-KD~\cite{adkd} and ReAugKD~\cite{reaugkd},
respectively.
Compared with AD-KD,
SinKD reduces the performance gap between the student and the teacher over 57\%,
highlighting 
that SinKD effectively narrows such gap by injecting structural knowledge from teacher to student.
Our improvements can be attributed to the unique properties of Sinkhorn distillation,
where the integrated characteristics of distributions are respected during distillation and thereafter facilitate impartial,
efficient knowledge transfer for robust convergence.
We also notice that SinKD does not rank the top on QNLI,
possibly due to suboptimal hyper-parameters for this specific task.
Meticulous tuning of hyper-parameters
might yield better results,
but will impair comparability and therefore is beyond the scope of the present study.

\subsection{Ablation Study}

\begin{table*}[htbp]
  \centering
  \small
    \begin{tabular}{l|ccccccc}
    \toprule
    \multirow{2}{*}{\textbf{Method}}  & \textbf{COLA} & \textbf{SST-2} & \textbf{MNLI} & \textbf{MRPC} & \textbf{RTE} & \textbf{QNLI} & \textbf{QQP} \\
    & (MCC) & (ACC) & (ACC) & (F1) & (ACC) & (ACC) & (ACC) \\
    \midrule
    SinKD (ours)  & 60.2 & 93.1 & 83.8/84.2 & 91.3 & 71.1 &90.5 &91.3 \\
    w/o $\mathcal{L}_{SD}$ & 53.6 & 91.1 & 82.7/83.1 & 89.4 & 66.8 & 90.1 & 90.5 \\
    w/o $\mathcal{L}_{KL}$ & 56.2 & 91.7 &82.3/83.0 & 90.1 & 69.3 &90.2 & 90.7 \\
    w/o $\mathcal{L}_{CE}$ & 58.0 & 92.3 & 83.5/84.1 & 91.1 & 70.4 &90.4 &91.3 \\
    w/o $\mathcal{L}_{KL}\&\mathcal{L}_{SD}$  & 51.2 & 91.0 & 81.7/82.6 & 89.2 & 66.1 & 89.3 & 90.4 \\
    \bottomrule
    \end{tabular}%
    
    \caption{Effect of different loss terms on GLUE.
    }
\label{ablation1}
\end{table*}%

\paragraph{Sinkhorn loss benefits the student the most among all losses.}
In order to study the impact of each loss component,
we carry out ablation studies on three variations of SinKD:
1) SinKD without Sinkhorn
loss,
2) SinKD without KL divergence loss,
and 3) SinKD without cross-entropy loss.
As revealed in Tab.~\ref{ablation1},
significant decreases over all tasks can be observed when Sinkhorn loss is removed.
In addition,
the drop of performance without KL divergence loss suggests that the proposed SinKD is supplementary to the vanilla KL divergence in distribution measurements.
With respect to the cross-entropy loss,
its supervision from ground-truth labels directly improves the student model and consequently should be kept intact during distillation.
Each component contributes to diminishing the gap between the student and the teacher.
Our proposed Sinkhorn loss brings the most pronounced gains over other losses,
confirming the validity of Sinkhorn distance as a stable metric for convergence to global optimum.

\begin{table*}[htbp]
  \centering
  \small
    \begin{tabular}{l|ccccccc}
    \toprule
    \multirow{2}{*}{\textbf{Level}}  & \textbf{COLA} & \textbf{SST-2} & \textbf{MNLI} & \textbf{MRPC} & \textbf{RTE} & \textbf{QNLI} & \textbf{QQP} \\ 
    & (MCC) & (ACC) & (ACC) & (F1) & (ACC) & (ACC) & (ACC) \\
    \midrule
    Sample-wise  &54.2&91.5&83.2/83.7&90.4&69.0&90.3	&91.2 \\
    Batch-wise & 60.2&93.1 &83.8/84.2 &91.3 &70.0 &90.5 &91.3 \\
    \bottomrule
    \end{tabular}%
    \caption{
    Comparison between the sample-wise and batch-wise SinKD on GLUE.
}
\label{ablation3}%
\end{table*}%

\paragraph{Batch-wise SinKD excels sample-wise SinKD.}
Tab.~\ref{ablation3} demonstrates the superiority of the batch-wise over the sample-wise SinKD on all tasks,
implying that the Sinkhorn distance is indeed adept in handling the deviation of the student from the teacher with a high-dimensional distribution.
The sample-wise distillation treats each instance independently while neglecting the overall tendency of the student in tracking distributions of the teacher.

\begin{table*}[htbp]
  \centering
  \small
    \begin{tabular}{l|cccccccc}
    \toprule
    \multirow{2}{*}{\textbf{Method}}  & \multirow{2}{*}{\textbf{Complexity}}& \textbf{COLA} & \textbf{SST-2} & \textbf{MNLI} & \textbf{MRPC} & \textbf{RTE} & \textbf{QNLI} & \textbf{QQP} \\ 
    & &(MCC) & (ACC) & (ACC) & (F1) & (ACC) & (ACC) & (ACC) \\
    \midrule
    RKL &$O\left(bd\right)$ &53.9 &91.6 & 82.9/83.4&90.5 &67.1 &90.1 &91.1 \\
    JS &$O\left(bd\right)$ &54.2 &92.2 &83.1/83.7 & 90.7&68.9 &90.3 &91.2 \\
    TVD &$O\left(bd\right)$ &54.1 &92.1 &83.3/83.8 &90.9 & 70.0&90.2 &91.2 \\
    \midrule
    SinKD &$O\left(b^2\left(d+T\right)\right)$ & 60.2&93.1 &83.8/84.2 &91.3 &71.1 &90.5 &91.3 \\
    \bottomrule
    \end{tabular}%
    \caption{ Comparison with distillation methods based on variants of $f$-divergence on GLUE.
}
\label{ablation2}%
\end{table*}%

\paragraph{SinKD surpasses distillation methods based on variants of $f$-divergence.}
To investigate if the existing distillation methods with $f$-divergence measures can achieve competitive results,
we replace our Sinkhorn loss with losses based on:
1) RKL divergence,
2) JD divergence,
and 3) total variation distance (TVD).
To fairly compare with SinKD,
each loss mentioned above is combined with cross-entropy loss and KL divergence loss during distillation.
Tab.~\ref{ablation2} shows that Sinkhorn distillation outperforms three other
distillation methods on all datasets,
verifying the superiority of Sinkhorn distance over variants of $f$-divergence measures in matching distributions.
Additionally,
it is worth noting that among the other three methods,
TVD exhibits
slight advantages over RKL and JS divergence on average.
Such finding is consistent with previous work~\cite{f-div}.


\begin{figure}[!ht]
\begin{center}
\includegraphics[width=\linewidth]{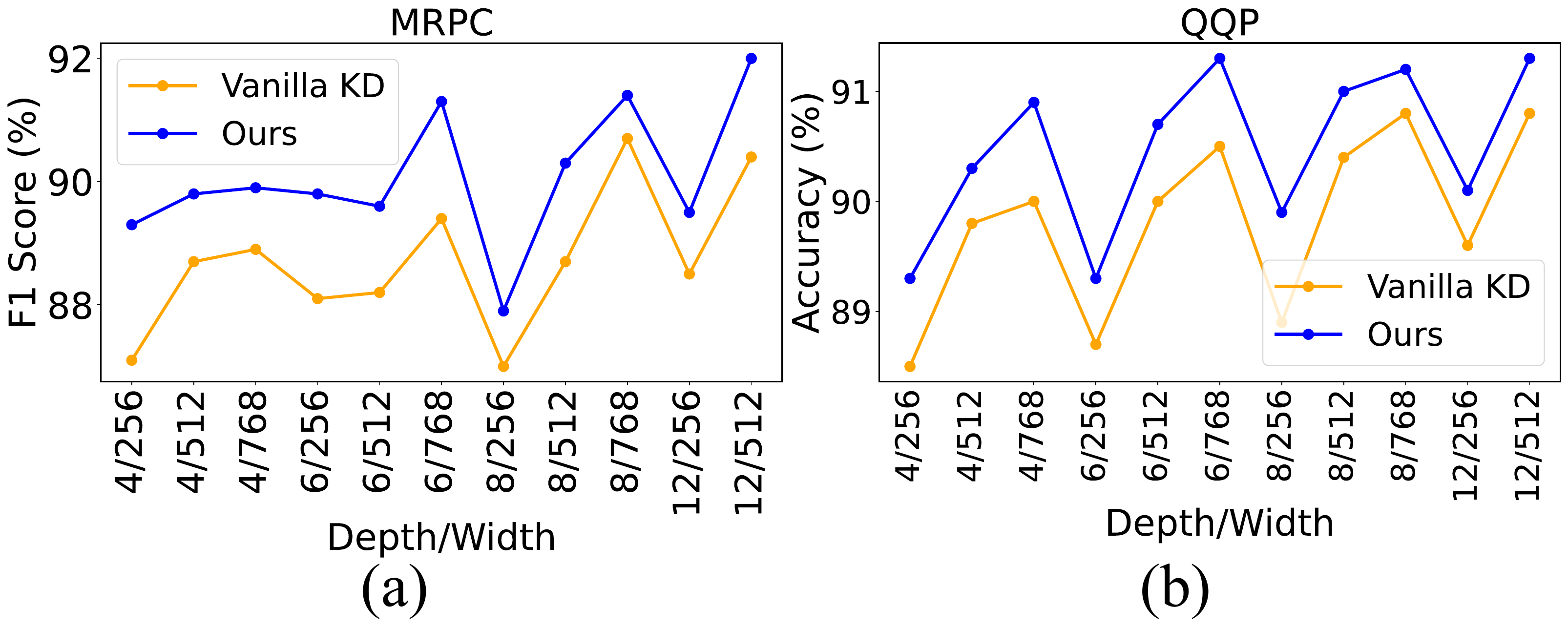}
\caption{
Performance at different student scales on (a) MRPC \& (b) QQP.
Best viewed magnified.
}
\label{scale}
\end{center}
\end{figure}

\paragraph{SinKD generalizes well on student LLMs across scales.}
To thoroughly assess the influence of size of student LLMs on the performance of SinKD,
we conduct an extensive analysis with comparison between the vanilla KD and SinKD.
Without loss of generality,
we take two tasks (MRPC and QQP) for demonstration.
A broad range of model scales~\cite{pd} are employed to explore the adaptability and robustness of SinKD when applied on student models with various configurations.
Note that both the vanilla KD and our SinKD are logits-based KD methods,
which are independent of model structure by nature and thus enjoy high versatility.
As illustrated in Fig.~\ref{scale},
SinKD consistently outperforms the vanilla KD on both two tasks across all scales. 
Such generalizability on model size
confirms the potential of SinKD as an efficient and reliable KD method.

\subsection{Discussion on Hyper-parameters}
\label{5.3}



\paragraph{$T$ as the number of Sinkhorn iterations}
We vary the number of iterations $T$ and results (see Tab.~\ref{ablation_temp})
reflect the importance of selecting an appropriate $T$.
An increase of $T$ to 20 respectively improves F1 scores for MRPC (91.3) and accuracy for QQP (91.3),
suggesting that
sufficient
iterations is crucial to approximation and
convergence.
Nevertheless,
raising the iterations to 50 yields no further improvement.
It indicates the existence of a saturation point,
beyond which additional iterations are not beneficial but redundant.
Hence,
we set $T=20$ throughout experiments.

\begin{table}[htbp]
  \centering
  \small
    \begin{tabular}{c|cc}
    \toprule
    \textbf{Number of}  & \textbf{MRPC} & \textbf{QQP} \\ 
    \textbf{Interations $T$} & (F1) & (ACC) \\
    \midrule
    2 &90.6 &91.0 \\
    5 &90.9 &91.0 \\
    10 &90.9 & 91.1 \\
    20& 91.3 & 91.3 \\
    50 & 91.3 & 91.3 \\
    \bottomrule
    \end{tabular}%
    \caption{
    Effect of $T$ on MRPC \& QQP.
}
\label{ablation_temp}%
\end{table}%


\begin{figure}[!ht]
\begin{center}
\includegraphics[width=1\linewidth]{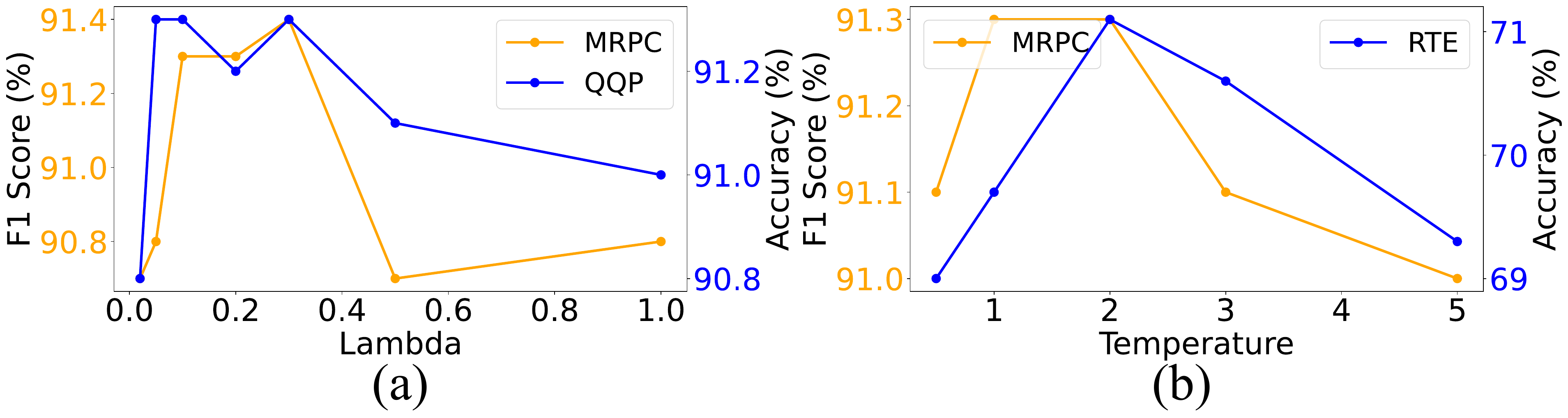}
\caption{
Effect of (a) $\lambda$ on MRPC \& QQP and 
(b) $\tau_{\text{SD}}$ on MRPC \& RTE.
Best viewed magnified.
}
\label{ablation4}
\end{center}
\end{figure}

\paragraph{$\lambda$ as the weight of entropy-regularization}
The Sinkhorn distance is derived from the entropy-regularized OT problem,
where the regularization term promotes a more dispersed and less concentrated OT plan.
In other words,
entropy-regularization would enhance the numerical stability and computational tractability of the solution to OT problem.
Theoretically,
$\lambda$ dictates the balance between the accuracy of the OT approximation and the stability of the solution.
A larger $\lambda$ results in a smoother and more stable solution,
albeit potentially less accurate.
A smaller $\lambda$ yields a more accurate solution at the risk of numerical instability.
As demonstrated in Fig.~\ref{ablation4}(a),
a $\lambda$ within the range of 0.1 to 0.3 appears to
achieve an optimal trade-off among various aspects.
Out of consistency,
we choose $\lambda=0.1$ throughout experiments.

\paragraph{$\tau_{\text{SD}}$ as the temperature in Sinkhorn loss}
Fig.~\ref{ablation4}(b) systematically investigates the influence of $\tau_{\text{SD}}$ on distillation on the tasks of MRPC and QQP.
Our findings indicate that the default empirical setting $\tau_{\text{SD}}=2$ is appropriate for both two tasks.
A smaller $\tau_{\text{SD}}$ may cause the student model to concentrate solely on learning the most salient features,
neglecting the nuanced but valuable information present in less probable categories for classification.
On the other hand,
a larger $\tau_{\text{SD}}$ results in smoother and more uniform probability distributions,
which 
confuses the student model to discern between
essential and irrelevant information.


\begin{figure}[!ht]
\begin{center}
\includegraphics[width=\linewidth]{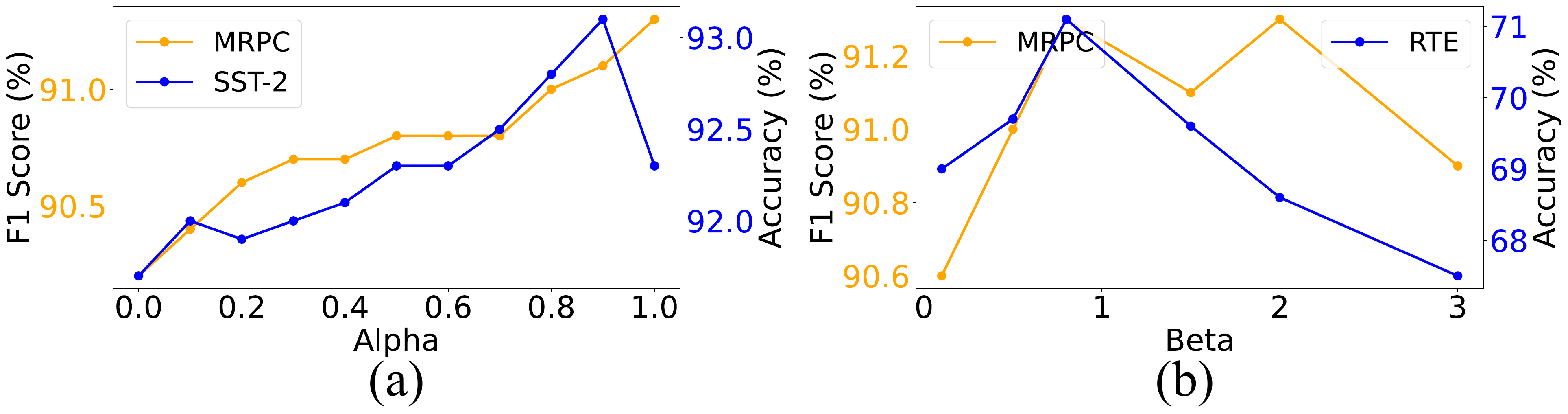}
\caption{Effect of (a) $\alpha$ on MRPC \& SST-2 and (b) $\beta$ on MRPC \& RTE.
Best viewed magnified.
}
\label{beta}
\end{center}
\end{figure}

\begin{table}[htbp]
  \centering
\small
    \begin{tabular}{c|cc}
    \toprule
    \multirow{2}{*}{\textbf{Batchsize $b$}}  & \textbf{MRPC} & \textbf{SST-2} \\ 
  & (F1) & (ACC) \\
    \midrule
    2 &90.5 &91.3 \\
    8 &90.8 &92.4 \\
    16 &91.3 & 92.8 \\
    32 & 91.1 & 93.1 \\
    64 & 91.3 & 93.1 \\
    \bottomrule
    \end{tabular}%

    \caption{
    Effect of $b$ on MRPC \& SST-2.
}
  \label{batch}%
\end{table}

\paragraph{$b$ as the number of batchsize}
In the present study,
the setting of batchsize is closely associated with the efficiency of geometric structural learning since the distribution divergences are measured within each batch of samples for the proposed Sinkhorn distance minimization.
An increased batch size is posited to enhance the student's understanding of complex geometric interrelations present within the dataset.
Empirical evidence,
as presented in Tab.~\ref{batch},
elucidates a positive correlation between augmented batch sizes and improved metrics (F1 scores for the MRPC benchmark and accuracy for SST-2).
Such performance gains are theoretically grounded in the premise that larger batches provide a more expansive dimensional space,
allowing for a more comprehensive representation of the geometric configuration during each optimization step.
A larger batch size $b$ effectively widens the model's exposure to the intrinsic geometric variance of the dataset,
potentially accelerating the transfer and assimilation of the teacher model's knowledge.
However, such benefit becomes negligible when the batch size increases beyond 32,
where both metrics for MRPC and SST-2 are almost unchanged.
This observation suggests the existence of a saturation point,
which delineates the boundary where the advantages of augmenting the geometric sampling space are outweighed by the computational overhead.

\begin{table}[htbp]
  \centering
\small
    \begin{tabular}{c|cc}
    \toprule
    \multirow{2}{*}{\textbf{Temperature $\tau_{\text{KL}}$}}  & \textbf{MRPC} & \textbf{SST-2} \\ 
  & (F1) & (ACC) \\
    \midrule
    1 &90.5 & 92.6 \\
    2 &90.8 & 93.1 \\
    3& 91.1 & 92.7 \\
    4 & 91.3 & 92.5 \\
    \bottomrule
    \end{tabular}%
 
    \caption{
    Effect of $\tau_{KL}$ on MRPC \& SST-2. 
}
 \label{temperature2}%
\end{table}

\paragraph{$\tau_{\text{KL}}$ as the temperature in KL loss}

Tab.~\ref{temperature2} provides the results of how the temperature $\tau_{\text{KL}}$ affects the knowledge distillation.
For the MRPC dataset,
a monotonically increasing trend in the F1-score is observed as $\tau_{\text{KL}}$ ranges from 1 to 4.
The best results of F1-score are achieved at $\tau_{\text{KL}} = 4$.
Conversely,
the accuracy for SST-2 is maximized at a lower temperature ($\tau_{\text{KL}} = 2$),
beyond which a diminution occurs.
It exemplifies the dualistic role of $\tau_{\text{KL}}$:
1) refining the granularity of probability distributions at lower temperatures and 2) fostering generalization at higher settings.
The optimal value of $\tau_{\text{KL}}$ is to be task-dependent,
underscoring the necessity for task-specific hyperparameter tuning in our SinKD applications.


\paragraph{$\alpha$ and $\beta$ as the loss weights}
In the total training objectives
of SinKD,
we introduce $\alpha$ and $\beta$ to
balance the contributions from the cross-entropy loss,
KL divergence loss,
and Sinkhorn distance loss.
A comprehensive evaluation of various combinations of $\alpha$ and $\beta$ can be found in
Fig.~\ref{beta}.
Each time,
we only adjust one parameter and keep the other one fixed.
Our findings indicate that a larger $\alpha$ 
generally produces better performance,
corroborating that knowledge transfer from the teacher model does play an indispensable role.
In line with the results of SinKD without the cross-entropy loss (see Tab.~\ref{ablation1}),
$\alpha=1$ causes a drastic decline on SST-2,
suggesting that ``soft" guidance from the teacher model is not equivalent to ``hard" supervision from ground-truth labels.
Additionally,
we observe that $\beta=0.8$ yields promising results for both two tasks.
Consequently,
we keep $\beta=0.8$ fixed and find the optimal $\alpha$ in \{0.8, 0.9, 1.0\} for each task.

\begin{table}[htbp]
\centering
\begin{minipage}{.47\linewidth}
\centering
\resizebox{\linewidth}{!}{
\setlength{\tabcolsep}{1mm}{
    \begin{tabular}{l|cc}
    \toprule
    \multirow{2}{*}{\textbf{Method}} & \textbf{RTE} & \textbf{CB} \\ 
                                     & (ACC)        & (ACC)       \\
    \midrule
    T0$_{\text{11B}}$ (T)            & 89.1         & 100         \\
    T0$_{\text{3B}}$ (S)             & 87.1         & 94.6        \\
    \midrule
    KL                               & 87.4         & 94.6        \\
    KL+RKL                           & 87.8         & 96.4        \\
    KL+JS                            & 88.1         & 96.4        \\
    KL+SinKD                         & \textbf{89.9}& \textbf{98.2} \\
    \bottomrule
    \end{tabular}
}
}
\caption{Results of T0 on SuperGLUE.}
\label{llm}
\end{minipage}%
\begin{minipage}{.5\linewidth}
\centering
\resizebox{\linewidth}{!}{
\setlength{\tabcolsep}{1mm}{
    \begin{tabular}{l|cc}
    \toprule
    \multirow{2}{*}{\textbf{Method}} & \textbf{RTE} & \textbf{CB} \\ 
                                     & (ACC)        & (ACC)       \\
    \midrule
    GPT$_{\text{1.3B}}$ (T)       & 75.4        & 86.9        \\
    GPT$_{\text{125M}}$ (S)       & 64.4         & 80.4        \\
    \midrule
    KL                               & 64.7         & 83.3        \\
    KL+RKL                           & 64.3         & 83.3        \\
    KL+JS                            & 64.6         & 82.1        \\
    KL+SinKD                         & \textbf{65.0}& \textbf{84.5} \\
    \bottomrule
    \end{tabular}
}
}
\caption{Results of GPT-Neo on SuperGLUE.}
\label{llm2}
\end{minipage}
\end{table}

\subsection{Generalizability on Generative LLMs}

To demonstrate the potential of our SinKD on generative LLMs,
we perform distillation on various transformer architectures including the encoder-decoder T0~\cite{t0} and the decoder-only GPT-Neo~\cite{gpt-neo}.
Specifically,
T0$_{\text{11B}}$ and GPT-Neo$_\text{1.3B}$ serve as the teacher while T0$_{\text{3B}}$ and GPT-Neo$_{\text{125M}}$ as the student.
We validate SinKD on the SuperGLUE~\cite{superglue} benchmark against SOTA KD methods based on
1) KL divergence,
2) RKL divergence,
and 3) JS divergence.
Note that we choose two datasets of RTE~\cite{rte} and CB~\cite{cb} for demonstrative experiments as they represent typical real-word NLP tasks.
Tab.~\ref{llm} and Tab.~\ref{llm2} show that the proposed SinKD surpasses all other KD methods.
Compared with the teacher GPT-Neo,
its student of 10 times fewer parameters can perform competitively with our SinKD.
Such findings showcase that SinKD can generalize to generative LLMs whose output logits are of high dimension equivalent to the size of the tokenizer vocabulary.
Moreover,
the performance gap between T0 and GPT-Neo can be ascribed to two reasons:
1) Architecture.
The encoder-decoder architectures are generally more suitable for discriminative tasks compared with the decoder-only architectures since the former better comprehend the input-output relationships with bi-directional modeling.
2) Model scale.
According to the scaling laws~\cite{gpt3},
the performance of GPT-Neo is expected to grow exponentially with billions of parameters increased.
Under the limited GPU budget,
experiments on larger decoder-only models are currently unavailable.


\section{Conclusion}
In this paper,
we resort to the Sinkhorn distance for divergence measure and present the SinKD to address the limitations of existing distillation methods.
Besides,
we propose a batch-wise reformulation to capture geometric intricacies of distributions across samples in the high-dimensional space.
Extensive experiments on the GLUE and SuperGLUE benchmarks confirm the superiority of our SinKD over SOTA methods from the aspect of comparability, validity, and generalizability.

A potential limitation is that we employ task formatting to adapt discriminative tasks under generative settings via prompts for experiments on GPT-Neo.
The manual design of these prompts requires engineering experience and could significantly influence performance.
Future work includes exploring application to representation-based KD and extension to other tasks (\textit{e.g.}, document summarization, machine translation).


\paragraph{Broader Impact} It is prospective to apply SinKD for distillation beyond the field of NLP.
Its advantage in handling the ``batchified" high-dimensional distributions would facilitate KD of the increasingly larger vision and language models for small-yet-competent ones with high cost-efficiency.


\section*{References}
\bibliographystyle{lrec_natbib}
\bibliography{reference}

\end{document}